%% file: acl_latex.tex
\pdfoutput=1

\documentclass[11pt]{article}

\usepackage[preprint]{acl}

\usepackage{times}
\usepackage{latexsym}

\usepackage[T1]{fontenc}

\usepackage[utf8]{inputenc}

\usepackage{microtype}

\usepackage{inconsolata}

\usepackage{graphicx}

\usepackage{amsmath}
\usepackage{amsthm}
\usepackage{amsfonts}
\usepackage{multirow}
\usepackage{booktabs}
\usepackage{pifont}
\newcommand{\cmark}{\ding{51}}
\newcommand{\xmark}{\ding{55}}

\title{CL-HOI: Cross-Level Human-Object Interaction Distillation from Vision Large Language Models}

\author{
  \textbf{Jianjun Gao\textsuperscript{1}},
  \textbf{Chen Cai\textsuperscript{1}},
  \textbf{Ruoyu Wang\textsuperscript{1}},
  \textbf{Wenyang Liu\textsuperscript{1}},
  \textbf{Kim-Hui Yap\textsuperscript{1}},
\\
  \textbf{Kratika Garg\textsuperscript{2}},
  \textbf{Boon-Siew Han\textsuperscript{2}}
\\
  \textsuperscript{1} School of Electrical and Electronic Engineering, Nanyang Technological University, Singapore \\
  \textsuperscript{2} Schaeffler Hub for Advanced Research at NTU, Singapore
\\
}

\begin{document}
\maketitle

\input{sec/0_abstract}
\input{sec/1_intro}
\input{sec/2_relatedwork}
\input{sec/3_method}
\input{sec/4_experiments}
\input{sec/5_conclusion}
\input{sec/6_limitation}
{
\bibliography{custom}
}

\input{sec/7_appendix}

\end{document}

%% file: sec/0_abstract.tex
\begin{abstract}

Human-object interaction (HOI) detection has seen advancements with Vision Language Models (VLMs), but these methods often depend on extensive manual annotations. Vision Large Language Models (VLLMs) can inherently recognize and reason about interactions at the image level but are computationally heavy and not designed for instance-level HOI detection. To overcome these limitations, we propose a Cross-Level HOI distillation (CL-HOI) framework, which distills instance-level HOIs from VLLMs’ image-level understanding without the need for manual annotations. Our approach involves two stages: \textbf{context distillation}, where a Visual Linguistic Translator (VLT) converts visual information into linguistic form, and \textbf{interaction distillation}, where an Interaction Cognition Network (ICN) reasons about spatial, visual, and context relations. We design contrastive distillation losses to transfer image-level context and interaction knowledge from the teacher to the student model, enabling instance-level HOI detection. Evaluations on HICO-DET and V-COCO datasets demonstrate that our CL-HOI surpasses existing weakly supervised methods and VLLM supervised methods, showing its efficacy in detecting HOIs without manual labels.

\end{abstract}

%% file: sec/1_intro.tex
\section{Introduction} 

\label{sec:intro} 

Human-object interaction (HOI) detection is a task to detect the human-object pairs and corresponding interactions in triplets of $<$human, interaction, object$>$. It is a fundamental task in computer vision and vital for surveillance~\cite{surveilliance}, human-robot interaction~\cite{hri}, and human-computer interaction \cite{hci}.

\begin{figure}[!ht] 

  \centering 

  \includegraphics[width=0.47\textwidth]{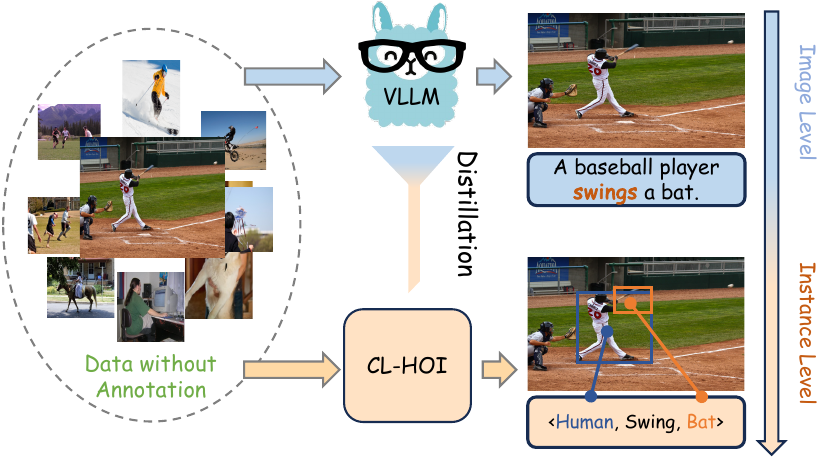}  

  \caption{\textbf{Cross-Level HOI Distillation}: VLLMs exhibit strong capabilities in image-level interaction recognition and reasoning. However, they are not equipped to perform instance-level HOI detection, which requires associating detected humans and objects. Our CL-HOI is proposed to distill image-level abilities to instance-level HOI detections.} 

  \vspace{-5pt} 

  \label{fig:motivation} 

\end{figure} 

HOI detection has increasingly adopted vision language models (VLMs) \cite{CLIP}, demonstrating significant potential in improving detection accuracy. Current HOI detectors primarily utilize these models to generate supplementary information or distill vision-text alignment, both of which enhance the learning process of HOI detectors. For generating supplementary information, some methods \cite{UniHOI} employ VLMs to produce descriptions of interactions or integrate them as sub-modules to improve detection performance, thereby providing additional context and understanding. For distilling vision-text alignment abilities, there are generally two approaches: prompting \cite{CLipHoi} and mimicking \cite{Gen-vlkt}. Prompting involves guiding vision encoders within VLMs to learn instance-level interaction features, effectively training the model to recognize and interpret specific interactions within an image. In contrast, mimicking aims to bridge the gap between the image-level features extracted by a customized vision encoder and those derived from the vision encoder in VLMs. Despite these advancements, these methods still rely on manual annotations at either the instance or image level.


Recently, vision large language models (VLLMs) \cite{BLIP, instructblip} have shown significant promise across various vision tasks, including image captioning and visual question answering. Trained on vast and diverse datasets, VLLMs exhibit remarkable zero-shot capabilities, enabling them to achieve image-level HOI understanding by recognizing human activities and reasoning about interactions, as illustrated in Fig. \ref{fig:motivation}. However, these capabilities are computationally intensive and remain confined to the image level, lacking the ability to perform instance-level HOI detection. Instance-level reasoning adds complexity and nuance by identifying and interpreting specific interactions between detected humans and objects, a task that current VLLMs are not fully explored. Even though RLIPv2 \cite{rlipv2} has adopted BLIP to generate captions, it uses some language tools and additional networks to generate interaction supervision, which is inefficient.


Building on the motivation presented in Fig. \ref{fig:motivation}, we introduce an annotation-free and Cross-Level HOI distillation (CL-HOI, student) framework to transfer the image-level HOI recognition and reasoning capabilities from the teacher VLLM \cite{instructblip} to instance-level HOI detection. The distillation is proposed in a two-stage manner: context distillation and interaction distillation.  
For \underline{context distillation}, we first prompt the teacher VLLM to generate context captions for each image, focusing on human activities. 
We then introduce a multimodal Visual Linguistic Translator (VLT) in CL-HOI that integrates a vision-language model with learnable adapters, enabling the translation of image features into context features and facilitating the learning of image-level human activities from context captions.
For \underline{interaction distillation}, we employ the language model (Vicuna \cite{vicuna}) within the VLLM to parse HOI triplets in the format $<$human, interaction, object$>$. 
Consequently, we introduce an Interaction Cognitive Network (ICN) in the student CL-HOI, which employs a novel chain-of-cognition strategy to effectively analyze HOIs through spatial, visual, and context cognition. This allows the model to learn instance-level interactions from parsed image-level HOI triplets, associating detected humans and objects.
To enable cross-level distillation, we introduce \underline{contrastive distillation losses} to capture contextual information in VLT and guide the ICN to detect interactions, thereby bridging the gap between image-level and instance-level detection.

Our contributions can be summarized as: (1) We introduce the pipeline that distills image-level knowledge from VLLMs to instance-level HOI detection without relying on manual annotation.
(2) We propose a new solution called CL-HOI, which includes context distillation and interaction distillation stages with the Visual Linguistic Translator and Interaction Cognition Network to effectively generate explicit interaction representations.
(3) We introduced contrastive distillation losses to facilitate the effective learning of context and interactions from VLLMs. We evaluated our methods on two benchmarking datasets. Our findings show that CL-HOI achieves state-of-the-art VLLMs supervised performance, with 17.5\% mAP on HICO-DET and 36.63\% Role AP on V-COCO.

%% file: sec/2_relatedwork.tex
\section{Related Work}
\vspace{-4pt}
\subsection{Human-Object Interaction Detection}
\vspace{-4pt}

Most existing HOI detection methods can be broadly categorized into groups, such as one-stage and two-stage methods.
Two-stage methods \cite{ican, Vsgnet, upt} typically apply an off-the-shelf object detector \cite{FRCNN, Detr} to detect the location of humans and objects, along with their semantic labels. Subsequently, detected boxes are fed to the interaction classification module for interaction modeling.
In addition to relying on object appearances through object detectors, the two-stage models can leverage contextual features \cite{ican,Vsgnet}, spatial
distribution \cite{SCG}, and human pose \cite{BodyHoi, PoseHoi} to enhance the analysis of 
Inspired by one-stage object detection \cite{Detr}, recent researchers attempted to develop end-to-end HOI detection \cite{DetrHoi, Ppdm, Hotr, ERNet} strategies for simultaneous detection of human/object localization and interaction classification. Some of the most recent methods \cite{MP-HOI, DP-HOI} aim to pre-train or train on extra relational data to improve performance.

Beyond one-stage and two-stage methods, which depend on expensive instance-level supervision, recent research has shifted towards weakly supervised methods. Kumaraswamy et al. \cite{MX-hoi} proposed a momentum-independent learning and element-swapping strategy for both fully and weakly supervised HOI detection. More recently, Unal et al. \cite{IO-HOI} introduced a method to detect HOIs based on image-level interaction labels (e.g., \textit{$<$jump$>$}, \textit{$<$hold$>$}), utilizing these labels for interaction detection.

\begin{figure*}[h!]
    \centering
\includegraphics[width=0.95\textwidth]{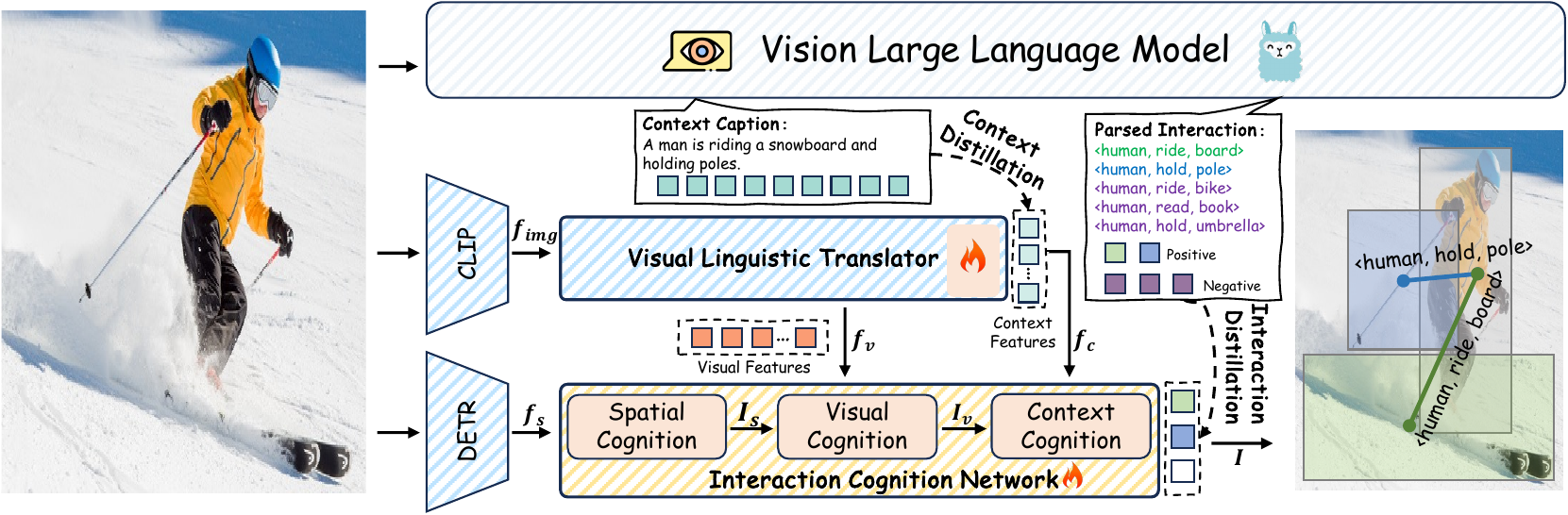}
    \caption{
    The proposed CL-HOI framework focuses on distilling instance-level interaction detection from the image-level recognition and reasoning abilities of the VLLM (teacher model). 
    Guided by context captions and parsed interaction triplets from the teacher model, CL-HOI performs context and interaction distillation for the Visual Linguistic Translator (VLT) and Interaction Cognitive Network (ICN) in two stages of instance-level interaction learning without annotated supervision.
    }
    \vspace{-4pt}
    \label{fig:method}
\end{figure*}

\subsection{Vision-Language Models in HOI Detection}

Many research works have investigated their application in incorporating various computer vision tasks with recent emerging developments in the vision language models. 
For instance, extensive research has been conducted on the popular vision-language models (e.g., CLIP \cite{CLIP}, BLIP \cite{BLIP}) in the context of HOI detection \cite{CLipHoi,viplo,Gen-vlkt}.
Gen-VLKT \cite{Gen-vlkt} introduces visual-linguistic knowledge transfer that relies on the semantic information extracted from a visual-linguistic pre-trained model to enhance interaction learning. 
Most existing multimodal-based HOI methods \cite{Gen-vlkt,CLipHoi} desire to distill the knowledge from large models and achieve better performance in supervised learning. More recent foundation models like VLLM and LLM have also been explored in HOI detections. Usually, such foundation models \cite{contexthoi, UniHOI} are prompted to provide additional cues to help detectors better learn HOIs.

\subsection{Knowledge Distillation}
Knowledge distillation \cite{KD} aims to enhance the learning capabilities of models by leveraging the strengths of large pre-trained models without relying on extensive labeled datasets. In knowledge distillation, a teacher model, usually a large and well-trained network, transfers its knowledge to a smaller student model, guiding the student to mimic the teacher’s predictions. Knowledge distillation eliminates the need for labeled data by using the intrinsic information within the data itself to guide the learning process. The teacher model generates pseudo-labels or other forms of supervision signals from the data, which the student model then uses to learn. These signals can come from various forms of supervision, such as predicting parts of the data from other parts \cite{ss-predicting}, identifying patterns within the data \cite{ss-representation}, or utilizing domain-specific transformations and augmentations \cite{ss-domain}.




%% file: sec/3_method.tex
\section{Methodology}
\label{sec:method}

\subsection{Overall Architecture}


The overall structure of our proposed CL-HOI framework is illustrated in Fig. \ref{fig:method}. This framework is designed to distill human-object interactions (HOIs) from a vision large language model (VLLM) through a two-stage process: context distillation and interaction distillation.
Within the student CL-HOI model, we initially use an off-the-shelf object detector (DETR) and a CLIP visual encoder to detect person-object pairs and extract spatial features \underline{$f_s$} as well as image features \underline{$f_{img}$}. The spatial features \underline{$f_s$} consist of human and object feature embeddings $e_h$ and $e_o$, bounding boxes $b_h$ and $b_o$, confidence scores $s_h$ and $s_o$, and object categories $c_o$, forming the person-object pairs $\{e_h, e_o, b_h, b_o, s_h, s_o, c_o\}$. 
In context distillation, we first prompt the VLLMs to generate the image captions (see details in Appendix). And then, we propose a Visual Linguistic Translator (VLT) to transfer the image features \underline{$f_{img}$} into context features \underline{$f_c$} and visual features \underline{$f_v$}, thereby distilling context captions from the VLLM. In the interaction distillation stage, we first use the LLM in the VLLM to parse the interactions. We also introduce a Chained Interaction Cognition Network (ICN) to reason about and prompt interactions based on spatial, visual, and contextual cognition, learning from the parsed interactions provided by the VLLM. Finally, we propose contrastive distillation losses to facilitate cross-level distillation and transfer the recognition and reasoning capabilities from the VLLM to the student model.

\subsection{Context Distillation Stage}




In the context distillation stage, we design a \underline{Visual Linguistic Translator (VLT)} to generate context features $f_c$ and visual features $f_v$ by translating the image features $f_{img}$ from visual space to linguistic context space, supervised by context captions prompted from the VLLM. Specifically, we insert two tuning adapters before and after the Q-Former \cite{BLIP}. Given the image features $f_{img}$ from the CLIP encoder \cite{CLIP}, the first adapter is used to map and project the image features to the Q-Former input space. Then, the projected feature with a set of initialized context queries $C_0$ are passed through the Q-Former to update the projected image features and context queries, obtaining updated image features $f'_{img}$ with context features $f_c$. Thus, the final context features can learn rich context information that is closely aligned with the image features. Besides the context feature generation, we introduce an additional tuning adapter to generate visual features $f_v$.  This can be achieved by aligning the image features $f'_{img}$ with the context features $f_c$ in the same dimension using an additional learnable projection layer.

Moreover, we propose a context distillation loss in Section \ref{sec:loss} to enable the bidirectional exchange between the visual and linguistic modalities within the proposed translator. Thus, it ensures that context and visual features mutually enhance each other, leading to a more accurate interpretation of human-object interaction reasoning in the subsequent phase.

\subsection{Interaction Distillation Stage}


In this stage, we introduce a chained \underline{Interaction Cognition Network (ICN)} that sequentially implements spatial, visual, and context cognitions to interpret and reason interactions from parsed interactions. The process begins with \underline{spatial cognition}, where geometric priors from detected person-object pairs are leveraged to generate initial interaction features with spatial information, denoted as $I_s$. This is followed by \underline{visual cognition}, which refines these features $I_s$ by incorporating detailed interaction features from the visual features $f_v$ and generates $I_v$. Finally, the \underline{context cognition} stage integrates context features $f_c$ into $I_v$ to obtain the interaction features enriched with contextual information, denoted as $I_c$. We utilize $I_c$ as the final interaction features $I$ for interaction learning. This chain of cognitive interaction reasoning creates a multi-layered interpretation of interactions, seamlessly blending geometric positioning, visual cues, and contextual narratives.
\vspace{3pt}
\noindent
\textbf{Spatial Cognition.}
To capture spatial relationships, we first generate a set of interaction features with spatial information $I_s$ from the detected spatial features $f_s$ represented as person-object pairs $\{e_h, e_o, b_h, b_o, s_h, s_o, c_o\}$. 
We derive spatial features based on relative positions $(c_h, c_o)$, dimensions $(w_h, h_h, w_o, h_o)$, areas $(a_h, a_o)$, aspect ratios $(a'_h, a'_o)$, intersection over union (IoU), and relative distances and directions $(d, d')$ from person-object pairs to capture spatial relationships.
These spatial features are projected into spatial priors $S$ using a linear layer. Simultaneously, content priors are refined through a self-attention mechanism within a Transformer encoder, modulating the embeddings $e_h$ and $e_o$ with their corresponding bounding boxes $b_h$ and $b_o$ by 
\begin{gather}
    E = {\rm SelfAttn}(e+{\rm PE}(b), e+{\rm PE}(b), e),
\end{gather}
where $PE(\cdot)$ is the positional encoding function. The spatial $S$ and content features $E$ are then concatenated to form a set of interaction features with spatial information $I_{s} = {\rm Concat}(E_h, E_o)$.

\noindent
\textbf{Visual Cognition.}
The visual cognition in the chain aims to refine $I_s$ to interaction features with visual information $I_v$ by integrating visual interaction features $f_v$ and guided by human-object union-region bounding boxes $b_u$.
This process initiates with generating $I_{v0}$ by augmenting $I_s$ with positional encodings $PE(c^x_u, c^y_u)$ \cite{pvic} specific to the center position $(c^x_u, c^y_u)$ of $b_u$. A multilayer perceptron (MLP) is proposed for dimensionality matching and feature enhancement. Concurrently, the visual features $f_v$ receive their positional encodings based on their spatial coordinates $(c^x_u, c^y_u)$, preparing them to serve as keys and values in the upcoming cross-attention operation. The cross-attention mechanism takes the prepared queries, keys, and values to produce updated interaction features $I_v$ imbued with rich visual interaction features. The process concludes with a pass through a feed-forward network (FFN), further refining these interaction features with visual information. The procedure to generate  $P_v$ can be formulated as:

\begin{small}
\begin{gather}
    I_{v0} = {\rm MLP}({{\rm Concat}(I_s, {\rm PE}(c^x_u, c^y_u))}),  \\
    I_v = {\rm FFN}({\rm CrossAttn}(I_{v0}, {\rm Concat}(f_v, {\rm PE}(c^x_u, c^y_u)), f_v)).
\end{gather}
\end{small}

\noindent
\textbf{Context Cognition.}
The context cognition phase within the cognition chain is crucial for obtaining interaction features within context information $f_c$ by weaving the context features $f_c$ to $I_c$. This process involves a cross-attention mechanism where the interaction features $I_v$ from visual cognition are used as queries, and the context features $f_c$ with rich context features serve as both keys and values. This attention-based interaction allows for the contextual and linguistic details to be aligned with and imposed on the interaction features, resulting in a new set of updated interaction features $I_c$. These are then fine-tuned through a feed-forward network (FFN), which refines the features and injects them with depth for a comprehensive understanding of interactions. The process to generate $I_c$ can be formulated as:

\begin{small}
\begin{gather}
    I_c = {\rm FFN}({\rm CrossAttn}(I_v, f_c, f_c)).
\end{gather}
\end{small}

For simplicity in subsequent discussions, we use $I$ to denote the final interaction features $I_c$, which are now thoroughly informed by contextual information, facilitating a more sophisticated interpretation of human-object interactions. Based on the final interaction features, we proposed interaction distillation losses in Section \ref{sec:loss} to achieve the cross-level distillation between the image-level supervision and instance-level interaction features.

\subsection{Contrastive Distillation Losses}
\label{sec:loss}

In the CL-HOI framework, contrastive distillation losses are essential for enabling information flow from the image level to the instance level. These losses are designed to leverage global similarities between features without requiring explicit matching between individual instances, drawing inspiration from DetCLIPv2 \cite{detclipv2}. 
This approach enables the model to learn from the overall relationships between features, facilitating the transfer of knowledge from broader image-level representations to more specific instance-level features, even when the exact correspondence between instances is unknown. Given any two features $f_1\in \mathbb{R}^{M \times D}$ and $f_2 \in \mathbb{R}^{N \times D}$, the similarities can be calculated as:

\begin{small}
\begin{gather}
s(f_1,f_2) = \frac{1}{M} \sum_{m \in M} \sum_{n \in N} \left( \frac{\exp(\operatorname{sim}_{m,n})}{\sum_{k \in N} \exp(\operatorname{sim}_{m,k})} \right) \operatorname{sim}_{m,n},\\
\operatorname{sim}_{m,n} = \langle f_1[m,\cdot], f_2[n,\cdot] \rangle, \quad \forall m \in M, \, n \in N.
\end{gather}
\end{small}

\noindent
Based on the similarity, we design context and interaction distillation losses to guide the two-stage distillation.
\vspace{5pt}

\noindent
\textbf{Context Distillation Loss.}
This loss is designed to optimize the context features $f_c$ to align with the context captions. It contrasts the similarity between the encoded context features $f_c$ and the caption features derived from the VLLM. In detail, generated captions in a mini-batch with $B$ captions are first encoded into caption features $f_{cap}$ using the text encoder of CLIP. Then, the context and caption features are then paired as $\{f^i_c, f^i_{cap}\}^B_{i=1}$. Then, we calculate the contrastive caption loss as:

\begin{small}
\begin{gather}
    L_c = -\frac{1}{B} \sum_{i \in B} \log{\frac{\exp{(s(f^i_c, f^i_{cap}))}}{\sum_{j \in B}{\exp{(s(f^i_c, f^j_{cap}))}}}}.
\end{gather}
\end{small}

\noindent
\textbf{Interaction Distillation Loss.}
In annotation-free settings, the absence of one-to-one instance-level matching necessitates carefully designing interaction distillation losses to improve HOI detection. We introduce three types of losses: 1) Image-to-Text (I2T) loss, 2) Text-to-Image (T2I) loss, and 3) Soft-Relation (SR) loss.
For interaction features $I$ in a mini-batch, we first encode verbs generated from the VLLM for each image into $f_{verb}$ using the text encoder in CLIP.
These verb features $f_{verb}$ paired with interaction features are treated as positives $f_{pos}$.
For negatives $f_{neg}$, we randomly sample negative verbs that are different from positive verbs from a verb dictionary (non-overlaps with ground-truth verb labels in original datasets) and encode them using the text encoder in CLIP. Further details on this process and the construction of the verb dictionary are provided in the Supplementary Materials. Thus, the interaction features and verb features in a mini-batch can be organized into $\{I^i \in \mathbb{R}^{K \times D}, f^i_{pos} \in \mathbb{R}^{P \times D}, f^i_{neg} \in \mathbb{R}^{Q \times D}\}^B_{i=1}$. 


\text{I2T loss} aims to ensure the learned interactions are tightly associated with positive verbs from image-level aspects. Specifically, the aggregated image-level interaction features are encouraged to align with positive verbs while contrasting against negative verbs. For each $I^i \in \mathbb{R}^{K \times D}$ in a mini-batch, we first average instance-level interaction features into image-level interaction features $I^i_{img} \in \mathbb{R}^{1 \times D}$. We then calculate the image-to-text loss as:

\begin{tiny}
\begin{gather}
    L_{I2T} = -\frac{1}{B} \sum_{i \in B} \log \frac{\exp(s(I^i_{img}, f^i_{pos}))}{\exp(s(I^i_{img}, f^i_{pos})) + \exp(s(I^i_{img}, f^i_{neg}))}.
\end{gather}
\end{tiny}

\text{T2I loss} is designed to ensure that positive verbs are represented in the learned interactions while negative verbs are excluded. The text-to-image loss is calculated as:

\begin{small}
\begin{gather}
    L_{T2I} = -\frac{1}{B} \sum_{i \in B} \log{\frac{\exp{(s(f^i_{pos}, I^i))}}{\sum_{j \in B}{\exp{(s(f^i_{pos}, I^j))}}}}.
\end{gather}
\end{small}

\text{SR loss} is inherited from weakly-supervised HOI detection, which uses CLIP \cite{CLIP} to generate pseudo-HOI labels from the generated interaction labels and cropped human-object regions. Details of the soft-relation loss $L_{SR}$ are provided in the supplementary materials.
The final loss is calculated as follows:

\begin{small}
\begin{gather}
    L = L_{c} + L_{I2T} + L_{T2I} + L_{SR}.
\end{gather}
\end{small}
\vspace{-10pt}

%% file: sec/4_experiments.tex
\section{Experiments}
\subsection{Setup}

\noindent
\textbf{Datasets. }
We evaluate the proposed method with well-established HOI detection datasets, HICO-DET \cite{hico-det} and V-COCO \cite{v-coco}.
HICO-DET consists of 37,633 and 9,546 images for training and testing, where the images are annotated with bounding boxes for human-object pairs and their interaction labels. This dataset includes 80 object categories (consistent with the MS COCO dataset \cite{v-coco}) and 117 interaction categories, resulting in a total of 600 unique $<$interaction, object$>$ pairs. 
The V-COCO dataset consists of 5,400 and 4,946 images for training and testing, respectively. It is a subset of the MS COCO dataset, which is annotated with 80 object classes and 26 interaction categories. Each image is associated with five captions.

\vspace{3pt}
\noindent
\textbf{Evaluation Metrics. }
We adopt commonly used evaluation metrics to evaluate the effectiveness of our proposed model, which uses Full mAP for HICO-DET and Role AP for V-COCO. 
It is worth noting that the evaluation matric of Full mAP and Role AP are analogous. Both metrics require that the predicted human and object bounding boxes exhibit an IoU of at least 0.5 with their respective HOI targets, and the predicted interaction category must align with the target interaction label.
\vspace{3pt}

\begin{table}[!ht]
\centering
\setlength{\tabcolsep}{3.2pt}
\renewcommand\arraystretch{1.1}
\resizebox{0.5\textwidth}{!}{
\begin{tabular}{l|c|c|c|c}
\bottomrule
Methods & Sup. & Ann. & Backbone & mAP \\
\hline
iCAN \cite{ican} & Full &\cmark  & RN50  & 14.84 \\ 
VSGNet \cite{Vsgnet} & Full &\cmark  & RN152 & 19.80 \\ 
SCG \cite{SCG}  & Full &\cmark  & RN50 FPN & 21.85 \\ 
IDN \cite{IDhoi}  & Full &\cmark  & RN50  & 23.36 \\ 
HOTR \cite{Hotr}  & Full &\cmark  & RN50+Tr. & 25.10 \\ 
MSTR \cite{mstr} & Full &\cmark  & RN50+Tr. & 31.17 \\ 
Gen-VLKT \cite{Gen-vlkt}   & Full &\cmark & RN50+Tr. & 33.75 \\ 
RLIPv2 \cite{rlipv2} & Full &\cmark &RN50+Tr. & 33.57 \\
PViC \cite{pvic} & Full &\cmark & RN50+Tr. & 33.80 \\
CMMP \cite{cmmp} & Full &\cmark & RN50+Tr & 33.24 \\
\hline
UPT (Baseline) \cite{upt}  & Full &\cmark  & RN50+Tr. & 31.66 \\
\textbf{CL-HOI (Ours)}  & Full &\cmark &RN50+Tr. & \textcolor{blue}{34.10} \\
\hline
MX-HOI \cite{MX-hoi} & Weak &\cmark & RN101 & \textcolor{blue}{16.14} \\
SCG \cite{SCG} & Weak &\cmark & RN50 FPN & 7.05 \\
IO-HOI \cite{IO-HOI}  & Weak &\cmark & RN50 FPN & 8.38 \\
\hline
RLIPv2 \cite{rlipv2} &VLLM &\xmark & RN50+Tr. & 15.05 \\
\textbf{CL-HOI (Ours)}  & VLLM &\xmark & RN50+Tr. & 15.62 \\
\textbf{CL-HOI (Ours)}  & VLLM &\xmark & Swin-L+Tr. & \textbf{17.55} \\
\toprule
\end{tabular}
}
\caption{Benchmark comparisons on the HICO-DET full mAP with both fully and weakly supervised settings. Sup: Supervision; Ann: Annotation; Tr: Transformer.}
\vspace{-10pt}
\label{tab: hico benchmark}
\end{table}

\begin{table}[!ht]
\centering
\setlength{\tabcolsep}{3.2pt}
\renewcommand\arraystretch{1.1}
\resizebox{0.5\textwidth}{!}{
\begin{tabular}{l|c|c|c|c}
\bottomrule

Methods & Sup. & Ann. & Backbone & Role AP \\
\hline
iCAN \cite{ican}  & Full &\cmark  & RN50  &  52.04 \\
VSGNet \cite{Vsgnet}  & Full &\cmark  & RN152 & 57.00  \\
SCG \cite{SCG}   & Full &\cmark  & RN50 FPN & 58.01  \\
IDN \cite{IDhoi}  & Full &\cmark  & RN50  & 60.30 \\
HOTR \cite{Hotr}   & Full &\cmark  & RN50+Tr. &  64.40  \\
MSTR \cite{mstr}  & Full &\cmark  & RN50+Tr. &  65.20 \\
Gen-VLKT \cite{Gen-vlkt}    & Full &\cmark & RN50+Tr. & 64.80 \\
PViC \cite{pvic}  & Full &\cmark & RN50+Tr. & \textcolor{blue}{67.80} \\
RLIPv2 \cite{rlipv2} &Full &\cmark &RN50+Tr. & 64.50 \\
CMMP \cite{cmmp} & Full &\cmark & RN50+Tr. & 61.20 \\
\hline
UPT (Baseline) \cite{upt} & Full &\cmark  & RN50+Tr. &  64.50 \\
\textbf{CL-HOI (Ours)}  & Full &\cmark &RN50+Tr. & 65.44 \\
\hline
SCG \cite{SCG}  & Weak &\cmark & RN50 FPN & 20.05\\
IO-HOI \cite{IO-HOI}   & Weak &\cmark & RN50 FPN & \textcolor{blue}{29.59} \\
\hline
\textbf{CL-HOI (Ours)}  & VLLM &\xmark & RN50+Tr. & 26.74 \\
\textbf{CL-HOI (Ours)}  & VLLM &\xmark & Swin-L+Tr. & \textbf{36.63} \\
\toprule
\end{tabular}
}
\caption{Benchmark comparisons on the V-COCO datasets with respect to Role AP with both fully and weakly supervised methods. Sup: Supervision; Ann: Annotation; Tr: Transformer.}

\label{tab: vcoco benchmark}
\end{table}

\noindent
\textbf{Baseline and Training Procedure. }
Our method is built upon a two-stage HOI detection method \cite{upt}. We utilize DETR \cite{Detr} that are pre-trained on MS COCO for object detection. We initialed the learning rate as $1e^{-4}$ and subsequently reduced it to $1e^{-5}$ after the 3rd epoch. The batch size is set to 16, and the model is trained for the five epochs. 


\begin{table}[t!]
\setlength{\tabcolsep}{3.2pt}
\renewcommand\arraystretch{1.3}
  \centering
    \resizebox{\columnwidth}{!}{
    \begin{tabular}{l|c|c|c}
    \bottomrule
    Method  & Annotation  & Backbone & Role AP \\
    \hline
    SCG \cite{SCG}   &\cmark & RN50 FPN & 20.05 \\ 
    IO-HOI (CC) \cite{IO-HOI}   &\cmark & RN50 FPN & 17.71 \\ 
    \hline
    CL-HOI (Ours)  &\xmark & RN50+Tr. & 26.74 \\
    CL-HOI (Ours)  &\xmark & RN50+Tr. & \textbf{30.07} \\
    \toprule
    \end{tabular}%
    }
    \caption{The comparison of our CL-HOI with weakly-supervised methods. Different from Table. \ref{tab: vcoco benchmark}, we also evaluate our CL-HOI with well-annotated captions from V-COCO.}
    \vspace{-15pt}
  \label{tab:v-coco-sup}%
\end{table}

\noindent
\subsection{Comparison to State-of-the-Art Methods}

The effectiveness of our proposed method is benchmarked through comparative analysis with current state-of-the-art (SOTA) methods in fully and weakly supervised Human-Object Interaction (HOI) detection. As shown in Table \ref{tab: hico benchmark} and Table \ref{tab: vcoco benchmark}, our proposed CL-HOI outperforms existing weakly-supervised methods on the HICO-DET and V-COCO datasets without the need for manual annotations. On the HICO-DET dataset, our CL-HOI, when using Swin-L as the backbone, achieves the best performance among weakly-supervised methods, reaching 17.55\% mAP. On the V-COCO dataset, CL-HOI with Swin-L outperforms all weakly-supervised methods, achieving 36.63\% Role AP. Even when adopting ResNet-50 as the backbone, CL-HOI demonstrates competitive performance compared to existing weakly-supervised methods on both HICO-DET and V-COCO datasets. For fully supervised comparisons, our CL-HOI shows performance comparable to existing methods. Notably, CL-HOI outperforms the baseline methods on both HICO-DET and V-COCO datasets. Overall, CL-HOI excels in HOI detection with minimal supervision, surpassing even some fully supervised counterparts.

\subsection{Weakly-Supervised CL-HOI}

To robustly demonstrate the effectiveness of our CL-HOI framework, we conduct a comprehensive evaluation on the V-COCO dataset under weakly-supervised settings, which is particularly notable for utilizing well-annotated captions. These captions serve as a rich source of supervision for our framework. The performance of CL-HOI under these settings is compared with other existing weakly-supervised methods and CL-HOI variants. This comparative analysis is detailed in Table \ref{tab:v-coco-sup}. As the comparisons indicate, our CL-HOI outperforms state-of-the-art methods under weakly-supervised settings. For a fair comparison, we include IO-HOI, which is trained using the well-annotated Conceptual Captions dataset \cite{conceptualconcept}. Moreover, even when trained on annotation-free captions, our CL-HOI continues to outperform existing methods.


%

\begin{table}[!t]
\renewcommand\arraystretch{1}
\setlength{\tabcolsep}{3.2pt}
\centering
\resizebox{\columnwidth}{!}{

\begin{tabular}{ccc|c|c}
\bottomrule
\multicolumn{3}{c|}{Prompting} & \multicolumn{1}{c|}{HICO-DET} & \multicolumn{1}{c}{V-COCO} \\ \cline{1-5} 
 Spatial  & Visual  & Context  & mAP & Role AP                         \\ \cline{1-5} \hline
    \xmark         & \xmark         & \xmark         & 9.42         &24.21        \\
    \cmark         & \xmark         & \xmark         & 11.24     &25.99         \\ 
    \cmark         & \cmark         & \xmark         & 14.67    &26.45   \\
    \cmark         & \cmark         & \cmark   & \textbf{15.62} & \textbf{26.74} \\ 

\toprule
\end{tabular}
}
\caption{Ablation study on the impact of chain-of-cognition cognition in interaction cognition network on HICO-DET and V-COCO datasets based on our baseline UPT.}
\vspace{-10pt}
\label{tab:ablation1}
\end{table}

\begin{figure}[h!]
    \centering
    \includegraphics[width=0.48\textwidth]{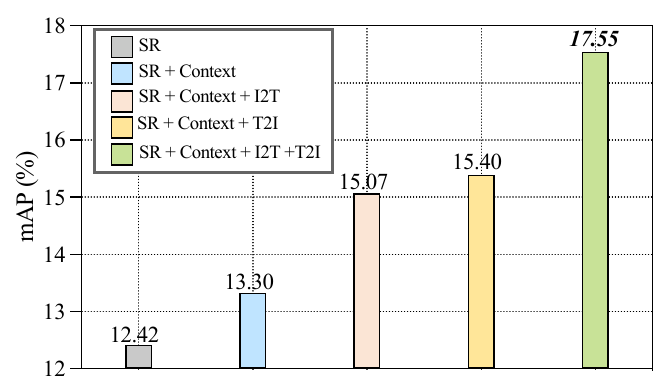}
    \caption{Ablation study on the impact of contrastive distillation losses on HICO-DET dataset.}
    \vspace{-10pt}
    \label{fig:contrastive}
\end{figure}

\subsection{Ablation Study}

\noindent
\textbf{Impact of  Different Teacher VLLM Models. }To evaluate the impact of different teacher models on HOI detection, we performed an ablation study using two VLLMs: BLIP-2 and InstructBLIP. As shown in Table \ref{tab: VLLM}, BLIP-2 with the RN50+Transformer backbone achieved an mAP of 9.63\%, which improved to 12.98 \% when using the more powerful Swin-L+Transformer backbone. InstructBLIP consistently outperformed BLIP-2, with an mAP of 15.62\% using RN50+Tr. and 17.55\% with Swin-L+Transformer. These results highlight the effectiveness of switching the teacher model from BLIP-2 to InstructBLIP, especially with stronger backbones.

\begin{small}
\begin{table}[!htp]
\renewcommand\arraystretch{0.9}
\setlength{\tabcolsep}{3.2pt}
    \centering
    \begin{tabular}{l|c|c}
    \bottomrule
            VLLM &          Bacbone &   mAP \\
    \hline
        \multirow{2}{*}{BLIP-2 \cite{BLIP}} &  RN50+Tr. &  9.63  \\ \cline{2-3}
          & Swin-L+Tr. & 12.98 \\
        \hline
        \multirow{2}{*}{InstructBLIP \cite{instructblip}} &   RN50+Tr. & 15.62 \\ \cline{2-3}
                                              & Swin-L+Tr. & 17.55 \\
    \toprule
    \end{tabular}
    \caption{Ablation study of different teacher models on the HICO-DET dataset. We evaluate two teacher models based on our CL-HOI.}
    \vspace{-10pt}
    \label{tab: VLLM}
\end{table}
\end{small}

\noindent
\textbf{Impact of the Chain-of-Cognition Strategy. }
First, we assess the impact of spatial cognition on HOI detection performance. Before evaluation, we use the baseline to train with the interaction distillation losses. Table \ref{tab:ablation1} shows that the baseline yields a mAP of 9.42\% and a Role AP of 24.21\% without our proposed VLT and ICN. We then add the spatial cognition to the baseline. The results show that spatial cognition improves the mAP by 1.82\% and the Role AP by 1.78\%, validating its effectiveness in providing crucial geometric information for HOI detection.
We then investigate the impact of visual cognition on the HOI detection performance. We add visual cognition to spatial cognition as the second chain. The results in Table \ref{tab:ablation1} show that visual cognition improves the mAP by 3.43\% and the Role AP by 0.46\%. The improvements demonstrate that the visual features are essential for HOI detection. It is worth noting that the VLT is not included in the above steps.
Lastly, we evaluate the impact of context cognition on CL-HOI. We introduce context cognition after visual cognition as the final chain. Table \ref{tab:ablation1} shows that context cognition improves the mAP to 15.62\% and the Role AP to 26.74\%. This confirms that contextual cues are also instrumental in accurately reasoning HOIs.

\vspace{3pt}
\noindent
\textbf{Impact of the Contrastive Distillation Losses. }
We also perform ablation studies on our proposed contrastive distillation losses in Fig. \ref{fig:contrastive} using Swin-L as the backbone. First, we applied only the soft-relation loss to the model, resulting in a mAP of 12.42\%. Next, we added the context loss, which improved the mAP to 13.30\%.  Following that, we incorporated the image-to-text and text-to-image losses, contributing additional improvements of 1.77\% and 2.10\%, respectively. Finally, when all the losses were combined, the model achieved a mAP of 17.55\%. These incremental improvements demonstrate that our proposed contrastive distillation losses significantly enhance the detection performance, making them a valuable component of our HOI detection methodology.

\noindent
\textbf{Random Initialization.}
We compare the performance of the trained model with a randomly initialized one on two benchmark datasets, as shown in Table \ref{tab:dataset_performance}. On HICO-DET, random initialization yields 5.30\%, improving to 15.62\% after training. Similarly, Role AP improves from 15.40 \% to 26.74\% with our framework. These results demonstrate the effectiveness of cross-level distillation in transferring image-level interaction reasoning to instance-level HOI detection.

\begin{table}[t!]
\renewcommand\arraystretch{1.2}
\setlength{\tabcolsep}{3.2pt}
\centering
\resizebox{0.5\textwidth}{!}{
\begin{tabular}{c|c|c|c|c}
\bottomrule
Datasets & Training & Backbone & mAP & Role AP \\ \hline
\multirow{2}{*}{HICO-DET} & w/o training & RN50+Tr. & 5.3  & -    \\ 
                          & w/ training   & RN50+Tr. & 15.62 & -    \\ \hline
\multirow{2}{*}{V-COCO}   & w/o training & RN50+Tr. & -    & 15.40 \\ 
                          & w/ training   & RN50+Tr. & -    & 26.74 \\ \toprule
\end{tabular}
}
\caption{Performance comparisons for HICO-DET and V-COCO datasets with training and without training (random initialization).}
\vspace{-5pt}
\label{tab:dataset_performance}
\end{table}

\subsection{Qualitative Analysis}

\begin{figure}[h!]
    \centering
    \includegraphics[width=0.48\textwidth]{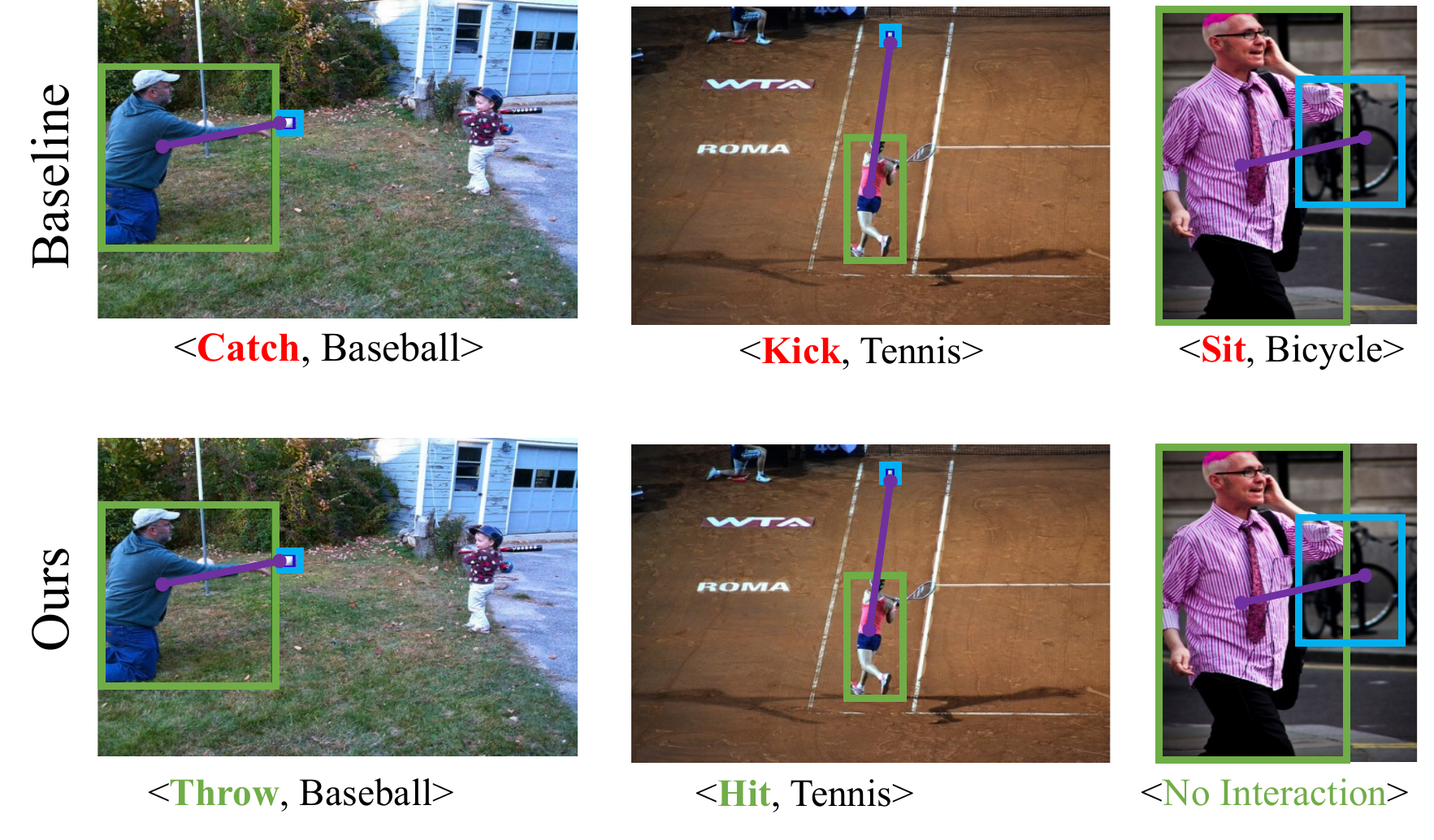}
    \caption{Qualitative comparisons of HOI detection performance between our baseline and CL-HOI.}
    \label{fig:qualitative}
    \vspace{-15pt}
\end{figure}

\noindent
The visualization in Fig. \ref{fig:qualitative} highlights CL-HOI's improved ability to detect human-object interactions compared to the baseline. In the first comparison, CL-HOI correctly identifies ``throw'', while the baseline mislabels it as ``catch''. Similarly, CL-HOI detects ``hit'' in the second group, which the baseline misses. In the third set, CL-HOI accurately identifies no interaction with a bicycle, unlike the baseline, which mistakenly detects ``sit''.

%% file: sec/5_conclusion.tex
\section{Conclusion}

We introduce CL-HOI, a human-object interaction detector that reduces reliance on manual annotations by distilling instance-level detection from VLLMs. Using a two-stage context and interaction distillation framework, we leverage a VLLM as a teacher model for cross-level knowledge transfer through a Visual Linguistic Translator (VLT) and Interaction Cognition Network (ICN). Contrastive distillation losses further enhance this transfer. CL-HOI achieves competitive results on V-COCO and HICO-DET, showcasing the effectiveness of our distillation approach.

%% file: sec/6_limitation.tex
\section{Limitations}
Our proposed CL-HOI aims to distill instance-level HOI detection from VLLMs, which are able to reason and recognize image-level human-object interactions. However, there are two bottlenecks for such cross-level distillation: the recognition accuracy of VLLMs and the relation ambiguities. The VLLMs fail to accurately recognize the image-level HOIs since they are not fine-tuned on HOI-related datasets. Thus, the image-level supervision are not as accurate as supervision in fully and weakly supervised methods. On the other hand, the image-level generated supervision is blind to the matching relations with the human-object pairs in the image. These two bottlenecks lead to an insufficient performance when evaluating on HICO-DET and V-COCO datasets.

%% file: sec/7_appendix.tex
\appendix
\goodbreak
\section{Appendix}

\subsection{Vision Large Language Model}

\begin{figure}[t!]
    \centering
    \includegraphics[width=0.49\textwidth]{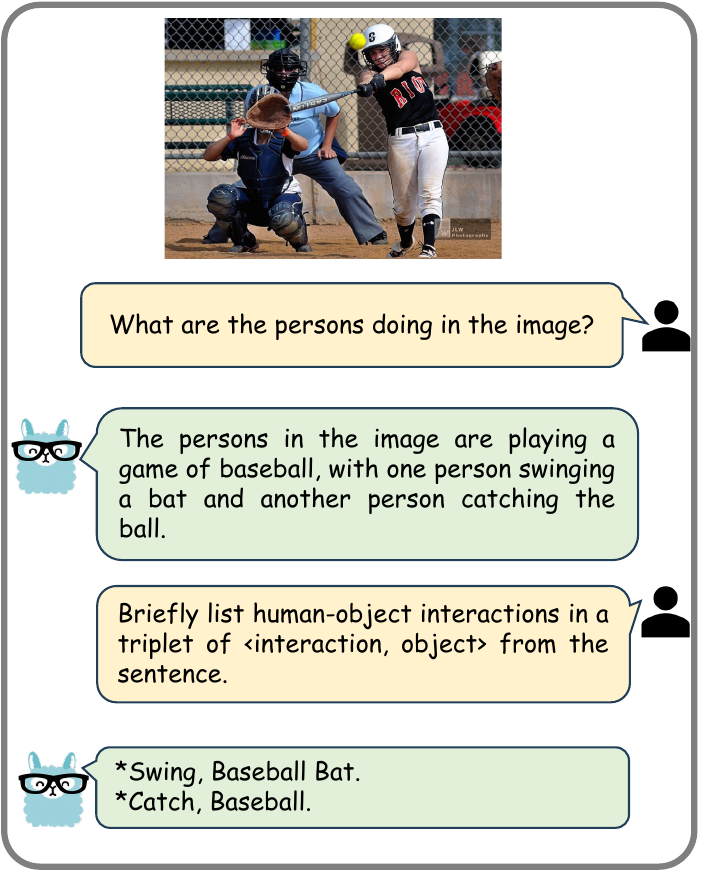}
    \caption{The example of the prompting implementation of the teacher VLLM.}
    \label{fig:vllm}
\end{figure}

In this paper, we introduce the InstructBlip \cite{instructblip} as the teacher model. In this section, we elaborate on the prompting implementation of the teacher model. 
We first prompt the teacher VLLM to generate the image-level context captions describing human activities. Given the input image, we use the IntructBlip ``What are the persons doing in the image?'' as $prompt_1$ to generate \textit{context caption}s. 
Based on the context captions, we further prompt the LLM (Vicuna \cite{vicuna}) in the InstructBlip to parse all possible human-object interactions. 
Specifically, we prompt ``Briefly list human-object interactions in a triplet of $<$interaction, object$>$ from the sentence.'' as $prompts_2$  to parse interactions from \textit{context caption}s. 
Then, we use the \textit{context caption}s and \textit{interaction triplet}s to supervise the learning of our proposed visual-linguistic translator and interaction cognition network. 
This process enables the context and interaction distillation and transfers the interaction recognition and reasoning capabilities from the VLLM to our proposed CL-HOI. We give an example to show how we generate the captions and parse the verbs in Fig. \ref{fig:vllm}. The final outputs of the VLLM are given in ``*Swing, Baseball Bat.'', ``*Catch, Baseball.''. We need some simple postprocessing to remove unnecessary punch marks.

\subsection{Negative Verb Generation}

\begin{figure}[!htbp]
    \centering
    \includegraphics[width=0.48\textwidth]{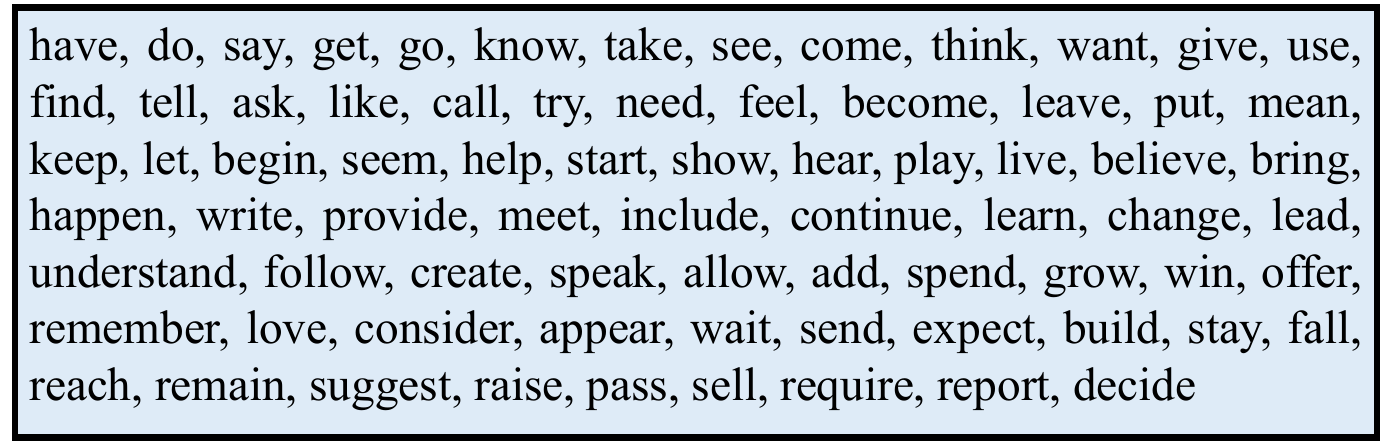}
    \caption{The verbs in the dictionary that are used to generate negative verbs in interaction distillation loss. For each valid image, we randomly choose two verbs from the dictionary as negatives.}
    \vspace{-10pt}
    \label{fig:dict}
\end{figure}

Providing the verbs parsed from the context captions, we identify these verbs as positive verbs. In cases where a caption lacks an action verb, we omit both the caption and its corresponding image from further consideration. This selective approach ensures that our analysis focuses only on images with captions that effectively describe actions. The negative verb generation process begins with constructing a verb dictionary, the contents of which are illustrated in Fig. \ref{fig:dict}. Notably, there is no overlap between the dictionary and ground truth verbs in both V-COCO and HICO-DET datasets. For each image, we randomly select two verbs from this dictionary that are not present in the positive verbs list, labeling these as negative verbs. Therefore, the data for training comprise context captions and positive and negative verbs for each input image.

\begin{figure*}[h!]
    \centering
    \includegraphics[width=\textwidth]{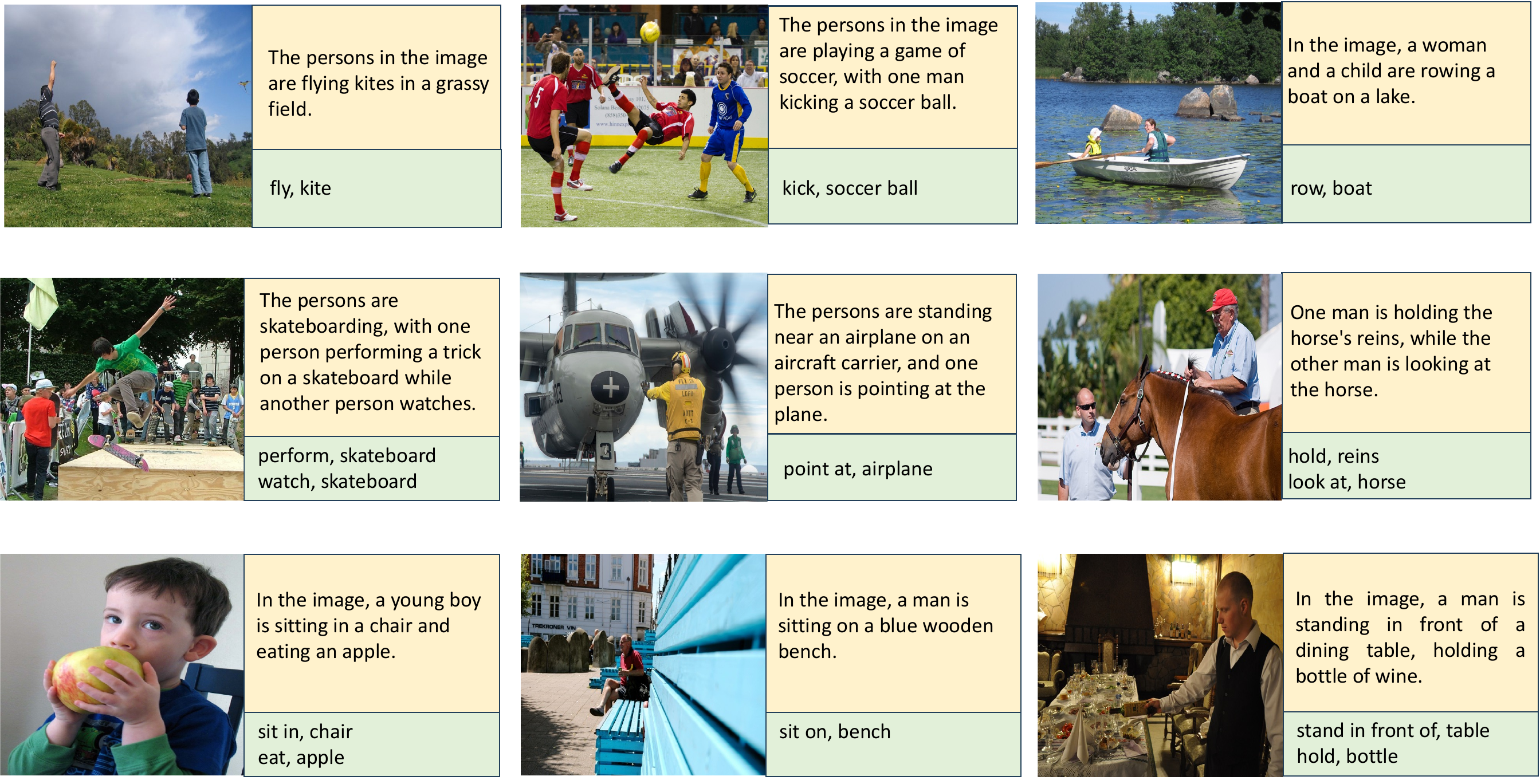}
    \caption{Examples of context captions and interactions generated from the teacher VLLM.}
    \label{fig:caption}
\end{figure*}

\begin{figure*}[h!]
    \centering
    \includegraphics[width=\textwidth]{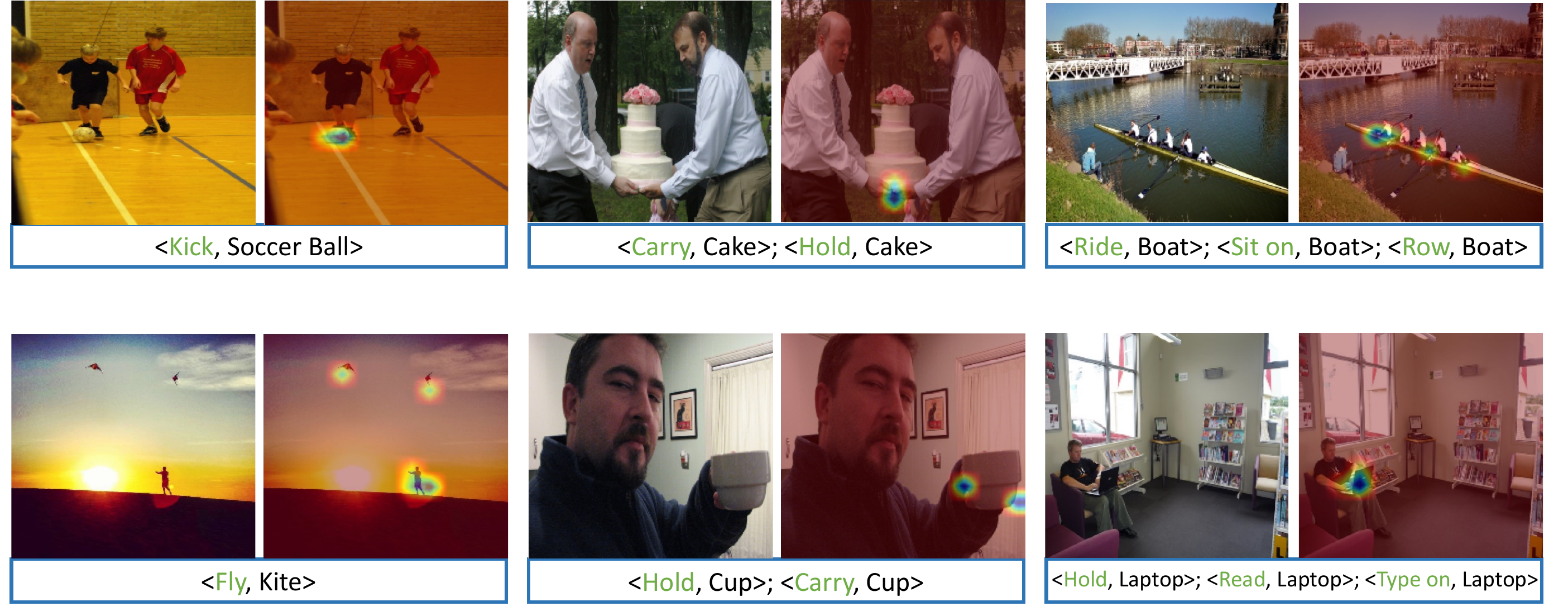}
    \caption{Visualized attention maps from CL-HOI, demonstrating the correctness and effectiveness of the context and interaction distillation. The six examples illustrate that CL-HOI accurately attends to the regions where interactions occur.}
    \label{fig:attn_s}
\end{figure*}

\subsection{Soft-Relation Loss}

In this paper, we also adopt the soft-relation loss in our interaction distillation loss. The soft-relation loss uses pseudo labels to supervise the interaction learning. Specifically, we first organize the detected human and object into union bounding boxes and the crop corresponding regions from images. After that, we use the CLIP vision encoder to encode these union regions as $f_u \in \mathbb{R}^{K \times D} $. On the other hand, we use the text encoder to encode positive verbs and objects from the parsed interaction triplets as $f_{pos} \in \mathbb{R}^{P \times D}$ and $f_{obj} \in \mathbb{R}^{P \times D}$. Then, we calculate the similarities between $f_u$ and $f_{pos}$ as well as $f_{obj}$ respectively as:

\begin{small}
\begin{gather}
sim_1 = \frac{f_u \cdot f^{\top}_{pos}}{\|f_u\| \|f_{pos}\| ^{\top}}, sim_2 = \frac{f_u \cdot f^{\top}_{obj}}{\|f_u\| \|f_{obj}\| ^{\top}}, 
\end{gather}
\end{small}

\noindent
where $\| \cdot \|$ denotes to row-wise $l_2$ norm, and $sim_1,sim_2 \in \mathbb{R}^{K \times P}$. The final pseudo-labels $Y_s$ are generated from these two similarities:

\begin{small}
\begin{gather}
    \tilde{sim}_{1}[k, p] = \begin{cases} 
    1 & \text{if } sim_{1}[k, p] \geq 0.5, \\
    0 & \text{if } sim_{1}[k, p] < 0.5,
    \end{cases} \quad \\
    \tilde{sim}_{2}[k, p] = \begin{cases} 
    1 & \text{if } sim_{2}[k, p] \geq 0.5, \\
    0 & \text{if } sim_{2}[k, p] < 0.5,
    \end{cases} \\
    Y_s = \tilde{sim}_{1} \odot \tilde{sim}_{2},
\end{gather}
\end{small}

\noindent
where $k \in K$ and $p \in P$, and $Y_s \in \mathbb{R} ^{K \times P}$.

Given the pseudo labels, we then calculate the soft-relation loss. Firstly, we calculate the similarities between interaction features $I$ and encoded positive verbs $f_{pos}$ as the final prediction as $\hat{Y}$ as 

\begin{gather}
    \hat{Y} = \frac{I \cdot f^{\top}_{pos}}{\|I\| \|f_{pos}\| ^{\top}}.
\end{gather}

The final soft-relation loss is calculated as follows:
\begin{small}
    \begin{gather}
    L_{sr} = - \sum_{k=1}^{K} \sum_{p=1}^{P} \left[ Y_{k,p} \log(\hat{Y}_{k,p}) + (1 - Y_{k,p}) \log(1 - \hat{Y}_{k,p}) \right].
\end{gather}
\end{small}

\subsection{Inference}

In the testing and inference phases of our approach,  we employ a method that centers around the use of interaction features, denoted as $I$, to predict the interaction class, represented as $y$. This is achieved through a process of calculating the similarity between $I$ and each of the encoded interaction labels within $Y$. The calculation can be presented as:

\begin{gather} 
    \hat{Y} = \frac{I \cdot {Y}^\top}{\|I\| \|Y\| ^{\top}} ,
\end{gather}
\begin{gather}
    s_i = {\rm sigmoid}(\hat{y}'_i)^{(1-\lambda)} \cdot (s_o \cdot s_h)^{\lambda},
\end{gather}
where $\lambda$  is a balancing factor that helps in appropriately weighing the contributions. 

\subsection{Label Generated from Teacher Model}
\begin{figure}[!htb]
    \centering
    \includegraphics[width=0.46\textwidth]{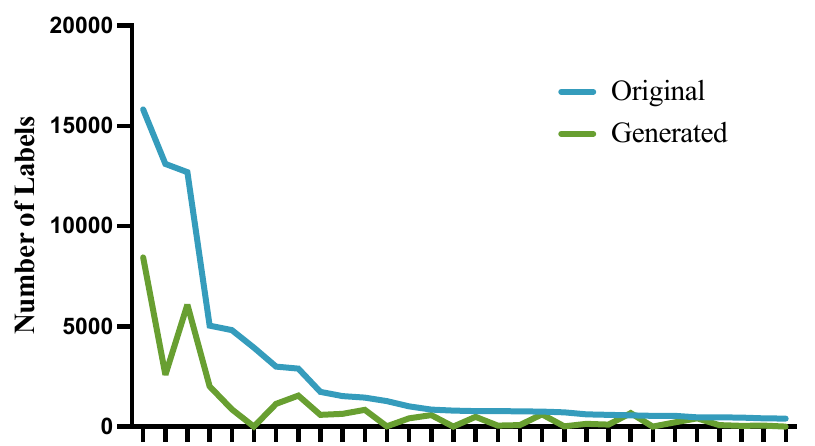}
    \caption{The quantitative comparisons of the top thirty labels with respect to numbers from the original HICO-DET dataset and labels generated from the VLLM.}
    \vspace{-10pt}
    \label{fig:data_analysis}
\end{figure}

Since the VLLM only provides image-level interaction labels, the grounding between weak labels and detected humans or objects is missing, unlike fully supervised settings. Additionally, the teacher model's supervision signals are incomplete, making the task harder than typical weakly supervised methods. We compare the labels from the original HICO-DET dataset with our VLLM-generated labels for the top 30 interaction categories in Fig. \ref{fig:data_analysis}. The results show a significant gap, particularly for categories with more instances, where original labels are nearly double our generated ones.

\subsection{Qualitative Analysis}

\textbf{Generated Context Captions and Interactions.} In Fig. \ref{fig:caption}, we show more examples of the context captions and interactions generated by teacher VLLM on the HICO-DET dataset. From the examples, it is evident that the VLLM is capable of recognizing image-level human activities and reasoning human-object interactions. For instance, the VLLM recognizes ``The persons in the first example are flying kites in a grassy field.'' and parses the ``fly, kite'' from the generated caption.

\vspace{8pt}
\noindent
\textbf{Attention Visualization.} We include more visualization results of the attention map of the last layer of VLT in Fig. \ref{fig:attn_s}. The results demonstrate that CL-HOI is able to capture context and interaction information to aid in attending to interactive areas, even when objects are not close to humans.